%% file: seeThrough.tex
\documentclass[10pt,twocolumn,letterpaper]{article}

\usepackage{cvpr}
\usepackage{times}
\usepackage{epsfig}
\usepackage{graphicx}
\usepackage{amsmath}
\usepackage{amssymb}
\usepackage{import}
\usepackage{soul}
\usepackage{subcaption}
\usepackage{rotating}
\usepackage{bm}
\usepackage{wrapfig}

\newcommand{\mypara}[1]{\vspace*{-0.1in} \paragraph{\bf #1.}}
\newcommand{\SeeThrough}{\mbox{\textsc{SeeThrough}}}
\newcommand{\seeingChairs}{\mbox{\textsc{SeeingChairs}}}
\newcommand{\fasterRCNN}{\mbox{\textsc{FasterRCNN3D}}}
\newcommand{\vnudge}{\vspace*{-0.1in}}

\newcommand{\AvgMaxIou}{\mbox{\textsc{IoU3D}}}
\newcommand{\AvgMaxDIoU}{\mbox{\textsc{IoU2D}}}
\newcommand{\PctCorrLoc}{\mbox{\textsc{Loc}}}
\newcommand{\PctCorrFull}{\mbox{\textsc{LocAng}}}
\newcommand{\AngDiff}{\mbox{\textsc{AngDiff}}}

\newcommand{\bb}[1]{{\bm{#1}}}

\usepackage[breaklinks=true,bookmarks=false]{hyperref}

\cvprfinalcopy % *** Uncomment this line for the final submission

\def\cvprPaperID{2338} % *** Enter the CVPR Paper ID here

\begin{document}

%%%%%%%%% TITLE
\title{\textsc{SeeThrough}: Finding Chairs in Heavily Occluded Indoor Scene Images}
%%under Heavy Occlusion using Scene %%Statistics}

%\author{submissionId: 2338}

%\if0
\author{Moos Hueting\\
University College London\\
{\tt\small m.hueting@ucl.ac.uk}
\and
Pradyumna Reddy\\
University College London\\
{\tt\small chinthala.reddy.17@ucl.ac.uk}
\and
Vladimir Kim\\
Adobe Systems\\
{\tt\small vokim@adobe.com}
\and
Nathan Carr\\
Adobe Systems\\
{\tt\small ncarr@adobe.com}
\and
Ersin Yumer\\
Adobe Systems\\
{\tt\small yumer@adobe.com}
\and
Niloy Mitra\\
University College London\\
{\tt\small n.mitra@ucl.ac.uk}
}
%\fi

\maketitle
%\thispagestyle{empty}

\input{abstract.tex}
\input{introduction.tex}

\input{relatedWork.tex}
\input{overview.tex}

\input{method.tex}

\input{results.tex}

\input{conclusion.tex}

{\small
\bibliographystyle{ieee}
\bibliography{seeThrough}
}

%\newpage
%\input{supplementary.tex}
%\input{appendix.tex}

\end{document}

%% file: abstract.tex
%%%%%%%%% ABSTRACT
\begin{abstract}

Discovering 3D arrangements of objects from single indoor images is important given its many applications including interior design, content creation, etc. 
Although heavily researched in the recent years, existing approaches break down under medium or heavy  occlusion as the core object detection module starts failing in absence of directly visible cues. 
Instead, we take into account holistic  contextual 3D information, exploiting the fact that objects in indoor scenes co-occur mostly in typical near-regular configurations.
First, we use a neural network trained on real indoor annotated images to extract 2D keypoints, and feed them to a 3D candidate object generation
stage. Then, we solve a global  selection problem among these 3D  candidates using
pairwise co-occurrence statistics discovered from a large 3D  scene database.
We iterate the process allowing for candidates with low keypoint
response to be incrementally detected based on the location of the already discovered nearby objects.
Focusing on chairs, we demonstrate significant performance improvement over combinations of state-of-the-art  methods, especially for scenes with moderately to severely occluded
objects. 
    
\end{abstract}

%% file: introduction.tex
% !TEX root =  seeThrough.tex

%%%%%%%%% BODY TEXT
\section{Introduction}

\begin{quote} \small 
    Partial occlusions pose a major challenge to the successful recognition of visual objects because they reduce the evidence available to the brain.  $\ldots$ As a result, recognition must rely not only on information about the physical object but also on information about the occlusion, scene context and perceptual experience~\cite{brainWorks17}.
\end{quote}

For many scene understanding tasks such as creating a room mockup for VR or automatically estimating how many people a room can accommodate, it is sufficient to estimate positions, orientations, and rough proportions of the objects rather than exact point-wise surface geometry. Given a {\em single} 2D photograph, the goal of this paper is to select and place instances of 3D models, particularly the partially {\em occluded} ones, to recover the photographed {\em scene arrangement} under the estimated camera.

%A quick glance at a single 2D indoor image is sufficient for humans to form a good idea about the 3D spatial arrangement of the room. Computationally mimicking the same is surprisingly challenging.  The problem is inherently ill-posed as photographs are the result of complex interaction between 3D geometry, material, illumination, optics, and the scattering of light. Given the high degree of complexity it is only natural to seek data-driven priors that can regularize the problem by leveraging 3D model and scene repositories.  In this paper, we investigate the following scenario: given a {\em single} 2D photograph, select and place instances of 3D  models to accurately recover the photographed {\em scene arrangement} under the estimated camera. 

\begin{figure}[b!]
\vnudge
    \includegraphics[width=\linewidth]{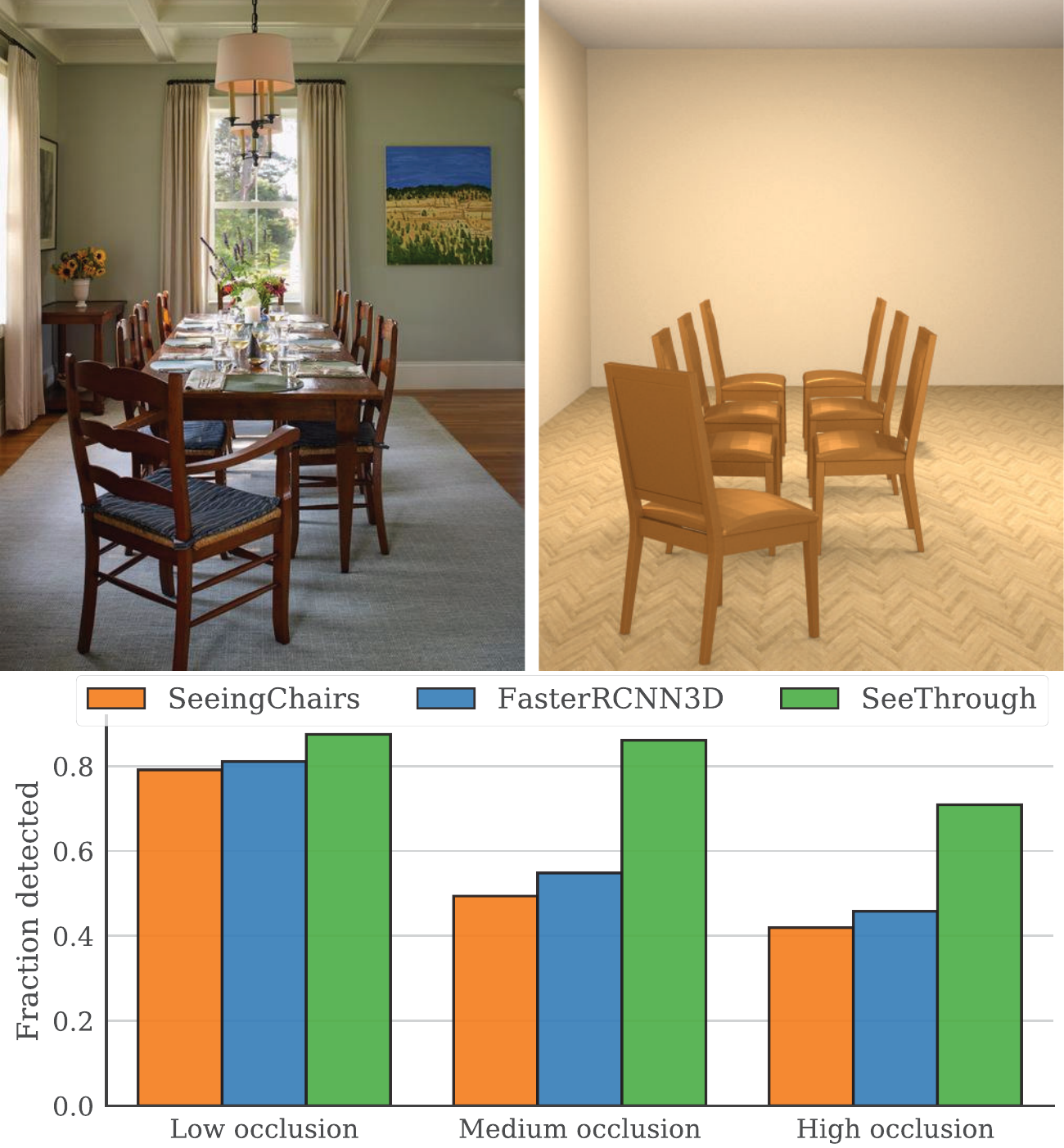}
    \caption[Context example]{We present \textsc{SeeThrough}, a method to detect objects (specifically chairs)  from single images under medium to heavy occlusion by reasoning with 3D scene-level context information. Our method significantly improves detection rate over state-of-the-art alternatives. }
    \label{fig:teaser}
\end{figure}

With the easy access to large volumes of image and 3D model repositories, and the availability of powerful supervised learning methods, researchers have investigated multiple subproblems relevant to the above goal, such as object recognition~\cite{He:2016:CVPR}, localization~\cite{Ren:2015:NIPS}, pose prediction~\cite{wu2016single}, or developed a complete system \textsc{Im2Cad}~\cite{izadinia2017im2cad} that selects and positions 3D CAD models that are similar to the input imaged scenes. While these approaches work reliably in rooms with relatively low occlusion, under moderate to heavy occlusion the methods quickly deteriorate. A common source of failure is that under significant occlusion, state-of-the-art semantic segmentation or region detection methods begin to break down, and hence any system relying on them also fail (see Figure~\ref{fig:teaser}).

Unlike images with limited occlusion where direct image-space information is sufficient, occluded scenes require a different treatment. 
One possibility is to train an end-to-end network to go from single images to parameterized scene mockups. However, a major bottleneck is obtaining suitable training data. On the one hand, in our experiments the networks trained with synthetic 3D scene data do not easily translate to real-world data. On the other hand, obtaining real-world training data is difficult to scale as it requires complex annotations in 3D from single images. Instead, we propose a novel approach that heavily relies on 3D contextual   statistics that can be automatically extracted from synthetic scene arrangement data.

Our key insight is that typical indoor scenes exhibit significant regularity in terms of co-occurrence of objects,
which can be exploited as explicit priors to make predictions about
object identity, placement and orientation, even under significant inter- or intra-object occlusions. For example, 
a human observer can easily spot heavily occluded chairs due to the presence of other visible nearby chairs and a table (see Figure~\ref{fig:teaser}), as we have a good mental model of typical chair-table arrangements. 

We introduce \SeeThrough\ that generates 2D keypoints from input images using a neural network, lifts the keypoints to candidate 3D object proposals, and then solves a selection problem to pick objects scored according to object cooccurrence statistics extracted from a scene database. We iterate the process by allowing already selected objects to reinforce selection of weakly witnessed occluded ones. 

We tested our approach quantitatively on a new scene mockup dataset including partially occluded objects and show significant improvement of recognition over baseline methods on multiple quantitative measures. Although our current
implementation is focused on the \emph{chair} class, the method itself is not
inherently limited to this, and could be extended to other classes with
appropriately annotated data.
{\em (Full code, training data, and scene statistics will be available for research use. Supplementary material is available at \url{http://geometry.cs.ucl.ac.uk/mhueting/proj/seethrough/seethrough_supplementary.tar.gz}) }

%% file: relatedWork.tex
%------------------------------------------------------------------------
\section{Related Work}
\label{sec:related}

\mypara{Scene mockups} 3D scene inference from 2D indoor images has recently received significant research focus due to the ubiquity of the new generation capture methods that enable partial 3D and/or depth capture. A significant amount of progress has been made following the early work of Hoeim et al.~\cite{hoiem2005automatic}, first with approximating only room shape~\cite{dasgupta2016delay,mallya2015learning,lim2014fpm,hedau2009recovering}, then inferring cuboid-like structures as surrogate furniture~\cite{del2012bayesian,choi2015indoor,zhang2014panocontext,xiao2012localizing,schwing2013box}. However, for detailed geometry prediction, the image input is generally supplemented with additional per pixel depth or point clouds~\cite{kmyg_acquireIndoor_sigga12}. Mattausch et al.~\cite{Mattausch:2014:CGF} used 3D point cloud input to identify repeated objects by clustering similar patches. Li et al.~\cite{Li:2015:CGF} utilize an RGB-D sensor to scan an environment in real time, and use the depth input to detect 3D objects queried from a database. While these works take 3D data as input, our method relies only on a single RGB image.

Recently, Izadinia et al.~\cite{Izadinia:2016:Arxiv} in their impressive \textsc{Im2CAD} system demonstrated
scene reconstruction with CAD models from a single image using image based
object detection (using FRCNN) and pose estimation approaches. Although their objective is similar to ours, the performance is bounded by the individual vision algorithms utilized in their pipeline. For example, if the segmentation misses an 
\begin{wrapfigure}{r}{0.43\columnwidth} 
\vspace{-10pt}
\hspace{-5pt}
  \includegraphics[width=0.43\columnwidth]{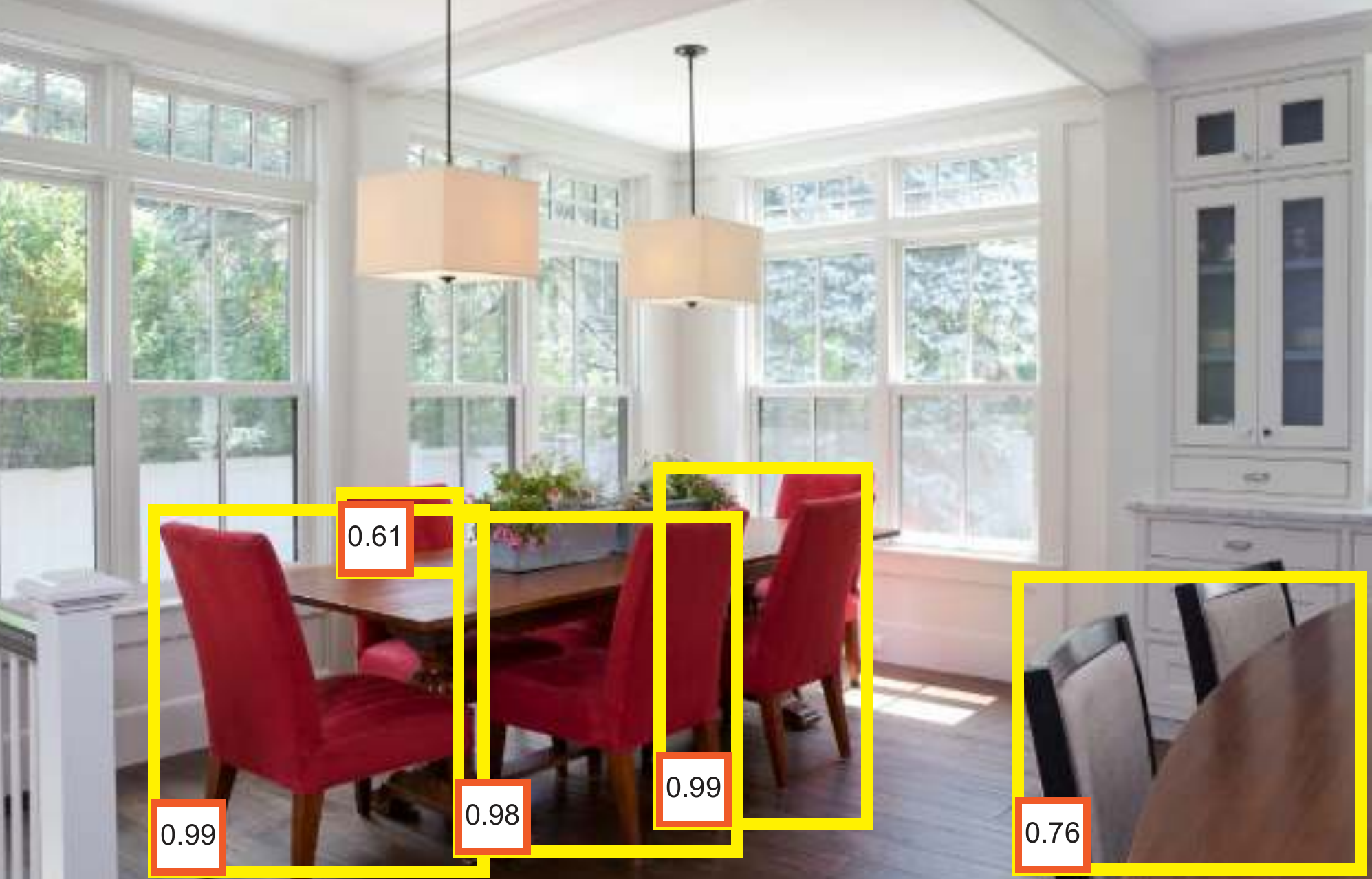} 
  \vspace{-20pt}
\end{wrapfigure} 
object because of significant occlusion (inset shows top FRCNN~\cite{Ren:2015:NIPS} detections with scores), there is no mechanism to recover it in the reconstruction (see Section~\ref{sec:ch4:evaluation} for comparison). On the contrary, our novel pairwise based search incorporates high level relationships typical to indoor scenes to recover from such failures successfully.

\mypara{3D$\rightarrow$2D alignment} Another way to create scene mockups is by directly fitting 3D models to the image. Pose estimation work~\cite{wu2016single,tulsiani2015viewpoints,huang2015single,lim2014fpm,kholgade20143d,Aubry:2014:CVPR} also demonstrated that given object images, reliable 3D orientation can be predicted, which in turn might help with scene mockups. Lin et al.~\cite{Lim:2013:ICCV} used local image statistics along with image-space features to align a given furniture model to an image. Aubry et al.~\cite{Aubry:2014:CVPR} utilized a discriminative visual element processing step for each shape in a 3D model database, which is then used to localize and align models to given 2D photographs of indoor scenes. Like most existing methods, their approach breaks down under moderate to high occlusion. Our method performs better, as other nearby objects can provide higher order information to fill in the lost information  (see Section~\ref{sec:ch4:evaluation}).

\mypara{Priors for scene  reconstruction} Scene arrangement priors have been successfully demonstrated in 3D reconstruction from unstructured 3D input, as well as scene synthesis~\cite{Fisher:2012:SIGGASIA}. Shao et al.~\cite{Shao:2014:SIGGRAPH} demonstrated that scenes with significant occlusion can be reconstructed from depth images by reasoning about the physical plausibility of object placements. Monszpart et al.~\cite{Monszpart:2015:SIGGRAPH} uses the insight that
planar patches in indoor scenes are often oriented in a sparse set of
directions to regularize the process of 3D reconstruction. On the other hand, based on priors between humans, Fisher et al.~\cite{Fisher:2015:SIGGRAPH} leveraged human activity priors together with object relationships as a foundation for 3D scene synthesis. In contrast to the complex and high order joint relationships used in these works, our object centric templates are compact and primarily encode the repetition of similar shapes (such as two side by side chairs) across pose and location.  This compact and simple template representation ensures that our search stays tractable at run-time. 

%% file: overview.tex
\section{Overview}
\label{sec:overview}

\begin{figure}[h!]
    \includegraphics[width=\linewidth]{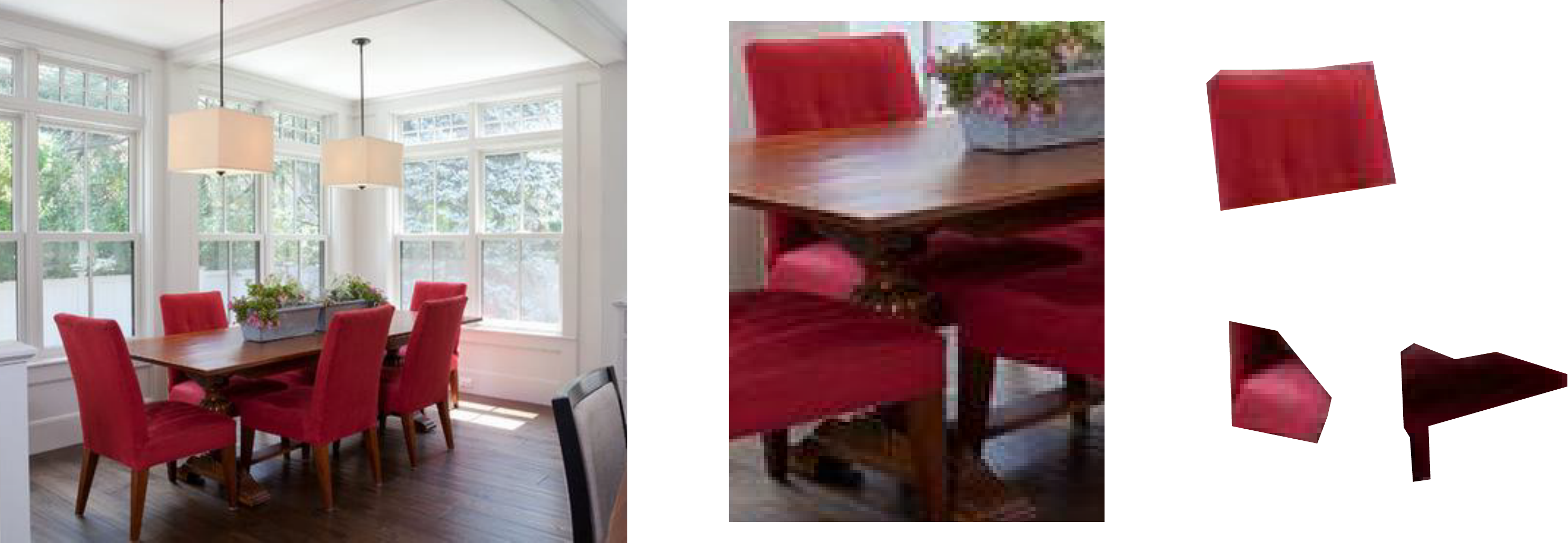}
    \caption[Decreasing context]{As humans, our understanding of scenes is heavily predicated on
    the context~\protect\cite{brainWorks17}. From left to right, less global information makes detection of chair harder.}
    \label{fig:context_example}
\end{figure}

In a scene with many chairs, we observe that the environment is not important for the recognition of the unoccluded chair  -- the shape of the object is clearly visible and
immediately recognizable. However, under  occlusion, the
task of recognizing the object necessitates adding 3D contextual information. State-of-the-art methods based on FRCNN~\cite{Ren:2015:NIPS} correctly detect chairs that are visible, but miss  partially occluded ones (see inset figure in Section~\ref{sec:related}). However, under  occlusion, the
task of recognition becomes easier with more contextual and cooccurence  information (see Figure~\ref{fig:context_example}).

Motivated by the above insight, we design  \textsc{SeeThrough} to run in three key steps: 
(i)~an image-space keypoint detection trained on AMT-annotated real photographs (Section~\ref{subsec:ch4:keypoint_maps}); 
(ii)~a candidate generation step that takes the estimated camera to lift detected 2D keypoints to 3D (deformable) model candidates  (Section~\ref{subsec:ch4:candidate_generation}); and 
(iii)~an iterative scene mockup stage where we solve a selection problem to extract a scene arrangement that proposes a plausible object layout using a common object co-occurrence prior  (Section~\ref{subsec:sceneInference}).

%% file: method.tex
% !TEX root = ./seeThrough.tex

\section{Method}

We now describe the three main steps of the \SeeThrough~system in detail starting with keypoint detection, followed by our approach for candidate object detection, and ending with our scene inference.

\subsection{Keypoint Detection}
\label{subsec:ch4:keypoint_maps}
At this stage our goal is to detect very subtle cues for potential object placements in a form of keypoints. A {\em keypoint} is a salient 3D point that appears across all objects of the same class (e.g., tip of a chair leg). We expect that a small number of (projected) keypoints will still be visible even under severe occlusions, and be useful in creating reasonable hypothesis for potential object placement. We represent this signal in two flavors: first, a \emph{keypoint map}, a per-pixel function that indicates how likely a particular keypoint is to occur at that pixel (each keypoint has a separate map $m_i$), and second, \emph{keypoint locations} which define the 2D coordinates for each keypoint. 
Both sets of information are used at different stages of our algorithm.  We collected our own training data and trained a convolution neural network to detect a continuous keypoint probability function, which we further use to extract candidate keypoint locations. 

%\textbf{Training Data.}
%We asked crowd-sourced workers to annotate 8 keypoints (Figure~\ref{fig:ch4:keypoint_types}) via a web-based interface (Figure~\ref{fig:ch4:amt}). 
%~\vk{some parts of text refer to 6 keypoints, probably a mistake} \vk{I suggest we move annotation interface to supplemental.}. 
%%
%\begin{figure}[h!tb]
%    \includegraphics[width=\linewidth]{figures/keypoint_types/keypoint_types}
%    \caption[Keypoint types]{Selected keypoint types.}
%    \label{fig:ch4:keypoint_types}
%\end{figure}
%%
%Our task included 500 images from the HOUZZ dataset and we recruited 3 workers per image via Amazon Mechanical Turk. We convolved these keypoints with Gaussian filter to make it easier for CNN to learn smooth filters and averaged the results~\vk{sigma?}. 
%%
%We also experimented with pre-training on synthetic data~\cite{Zhang:2017:CVPR} for training, but found that it does not improve network performance. 

%\textbf{Neural Network.}
We picked $N_k$ keypoints ($N_k=8$ in our tests)  (see supplemental material) and fine-tuned a variant of ResNet-50 neural network~\cite{He:2016:CVPR} to predict these keypoint maps in $N_k$ output channels (see supplemental material for architecture details). We also tested the CPM architecture~\cite{Wei:2016:CVPR}, but it yielded slightly inferior performance. While the latter focuses on keypoint detection it was pre-trained on human poses rather than general images, which is why we believe CPM did not generalized as well to our particuar task (see supplemental material). 

\begin{figure}[h!]
    \includegraphics[width=\linewidth]{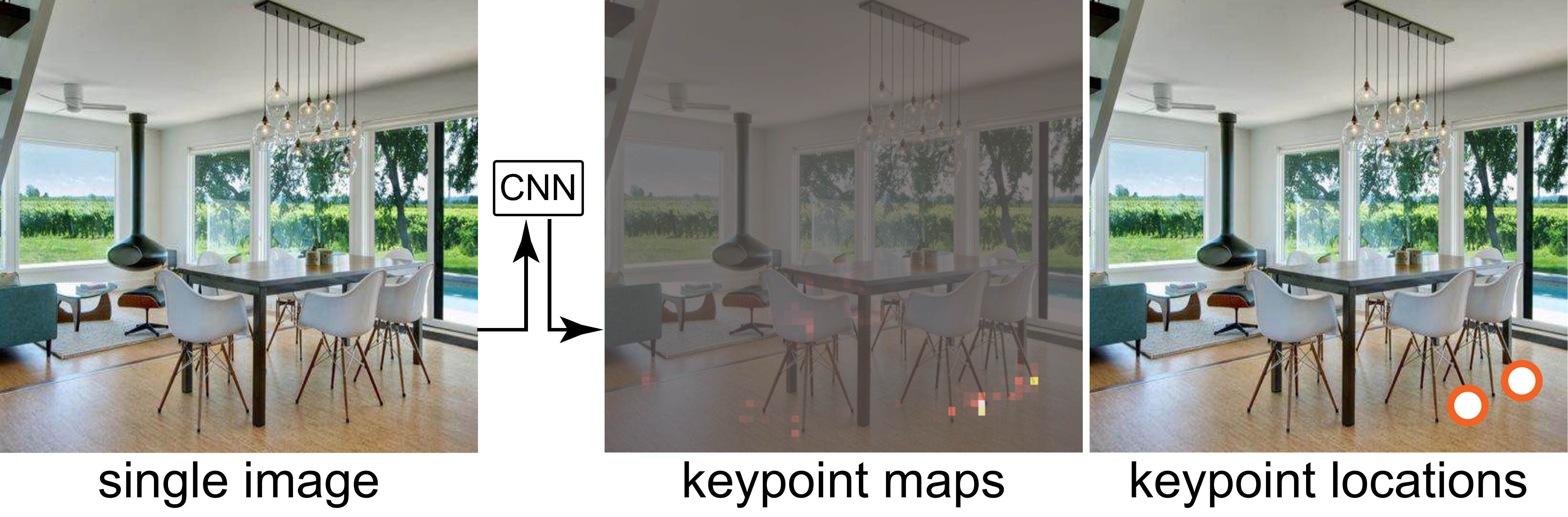}
    \caption{We trained a neural network on real images to detect {\em keypoint maps}, which are then converted to 2D {\em keypoint locations} via thresholding and non-maximal suppression. }
    \label{fig:ch4:keypoint_map_to_keypoints}
\end{figure}

The above network predicts continuous keypoint maps $\mathbb{M} := \{m_1, ..., m_{N_k}\}$, and to extract the final keypoint locations (2D positions in the image) we used local maxima above a threshold $\tau_m$ (Figure~\ref{fig:ch4:keypoint_map_to_keypoints}). We denote the set of these keypoint locations by $\mathbb{Q} := \{\bb{Q}_1, \ldots, \bb{Q}_{N_k}\}$. 

\subsection{Candidate Object Detection}
\label{subsec:ch4:candidate_generation}
The goal of this step is to propose multiple candidate objects based on the detected keypoints. While we do not know how to group points, we observe that 
a very small number of keypoints (as few as two) belonging to the same object, provide enough constraints to infer the scale and the orientation of a proxy 3D object. Hence, we can generate multiple candidates even with a sparse signal under moderate to high levels of occlusions. Using these generated candidates, we can recast the global inference problem as a discrete graph optimization problem, where we only need to solve for indicator variables, selecting a subset of candidates. Thus, we want higher recall at the expense of lower precision in this step. 
Furthermore, in order to incorporate a slightly bigger context than a single keypoint, we select subsets of points that can compose an object. 
At training time we learn a deformable template from a database of 3D models, and at test time we optimize the fitting of these templates to various subsets of keypoints.

\mypara{Object template} 
Given a database of consistenly aligned 3D models $\bb{M}$ with manually labeled keypoints we use Principal Component Analysis (PCA) to 
project 3D coordinates of keypoints to a lower-dimensional space (we take eigenvectors $\bb{\lambda}_1, ..., \bb{\lambda}_k$ that explain $>85\%$ of the variance). Our template is parameterized by a linear combination of these eigenvalues with weights $\bb{p}=[p_1, ..., p_k]$ (representing offset from the mean $\bb{\lambda}_0$). The final object template is defined by a weighted linear combination of the eigenvectors: $T(\bb{p}) := \bb{\lambda}_0 + \sum_i p_i \bb{\lambda}_i$.
%

%\subsubsection*{Object fitting} 
We formulate an optimization problem where we solve for object parameters (i.e., $\bb{p}$) while making sure that the object aligns with the detected keypoints. To relate our 3D deformable model to 2D images, we need a camera estimate. We use a variant of Hedau et al.~\cite{Hedau:2009:ICCV} to estimate a rotation matrix $C_R$ with respect to the ground plane, the focal length $C_f$, and define the camera's location $C_t$ to be at eye height (1.8m) above the world origin, giving camera parameters $C : =[C_R, C_f, C_t]$.
For each object we solve for a 2D translation across the ground plane $\bb{t}$, azimuth $\theta$, scale $s$, and 3D chair template parameters $\bb{p}$.  
Hence, the reprojection $z_i$ of the $i$-th keypoint to image space is:
\begin{align}
z_i := \Pi_C\left(R_\text{up}(\theta) ~ s ~ k_i(\bb{p}) + \bb{t}\right), 
\end{align}
where $k_i(\bb{p})=[T(\bb{p})]_i$ is a keypoint on the deformed template, $R_\text{up}$ is a rotation around the up vector,  and $\Pi_C$ is a projection to the camera space. 

As described next, we fit our template object in two stages: first, we propose a candidate based on a pair of points, and then, we refine these candidate parameters with respect to all keypoint maps. 

\mypara{(i)~Initial proposals} To propose initial object candidates we sample all pairs of detected keypoints. We use a pair because it gives the smallest set to sample that provides enough constraints to extract an initial guess for object translation, scale, and orientation.  For each pair, we initialize as $\bb{t} = \bb{0}, \theta = 0, s = 1, \bb{p} = \bb{0}$, and optimize:
\begin{align}
L_\text{init} = \sum_{i \in \{u, v\}} \|z_i - k_i \|^2 + \underbrace{
\alpha_1 \|s-1\|^2 + \alpha_2 \|p\|^2
}_{\text{regularizer } (L_\text{reg})}, 
\label{eqn:energy}
\end{align}
where $\alpha_1$ and $\alpha_2$ are respectively the weights balancing scale and deformable template parameters   ($\alpha_1 =1$ and $\alpha_2 = 1$ in our tests).

\mypara{(ii)~Parameter refinement} For each of the initial proposals extracted above, we refine the fitting. Specifically, instead of considering point-locations, we define our objective with respect to soft keypoint maps $m_j$, maximizing the probability of template corners to align with keypoints predicted by the neural network, i.e., 
\begin{align}
L = \sum_{i \in \{1, \ldots, N_k\}} \|1 - m_i(z_i)\|^2 + L_\text{reg}, 
\end{align}
with $L_\text{reg}$ as defined in Equation~\ref{eqn:energy}. 
If $L < \tau_u$, we add the final parameters as a candidate placement to our candidate placement set $\bb{O}$. 

\mypara{Selecting a 3D mesh} 
For the results presented in this paper we show 3D meshes rather than object templates. 
Particularly, we pick the closest 3D model from our database by projecting its keypoints
into the object PCA space, finding the nearest neighbor of the deformed template, and finally deforming it using the optimized parameters $\bb{p}$.

\subsection{Scene Inference}
\label{subsec:sceneInference}
We do not expect all individual objects selected as candidates to be in the scene, since they might overlap, or have inconsistent arrangement. 
First, we capture scene statistics obtained from a large scene dataset with a probabilistic model, and then use the model to formulate an alternating discrete and continuous optimization.

\mypara{Learning scene model}
We model higher level scene statistics via a graphical model where each object is a node and edges between pairs of nodes capture object-to-object co-occurrence relationships. We used a Gaussian Mixture Model (GMM) with $N_m$ (set to $5$ in our tests) mixture components to model relative orientation $\delta_\theta$ and translation $\bb{\delta}_t$ of pairs of chairs from a very large synthetic scene dataset~\cite{Zhang:2017:CVPR}. We only take into account chairs that
are within a distance $\delta_r = 1.5m$ from each other, reasoning that far-away objects have weaker relationships. 
We use Expectation-Maximization algorithm to fit the GMM and add a small bias (0.01) to the diagonal
of the fitted covariance matrices since objects in the database are axis-aligned.

\mypara{Graph optimization}
We formulate a graph labeling problem to decide which of the candidate objects should be included in the scene mockup, denoted by indicator variable $\gamma_i \in \{0,1\}$, where $\gamma_i=1$ iff object $O_i$ is included. We {\em minimize} the following objective function:
\begin{align}
L_\text{graph} := \sum_i \gamma_i U_i + \sum_{i,j} \gamma_i \gamma_j P_{i,j},  
\end{align}
where $U_i$ is a unary penalty for an included object, and $P_{i,j}$ is pairwise penalty for a pair of included objects. 

We define the unary energy by  projecting object's keypoints to the image and convolve the resulting keypoint map with a Gaussian, following the same procedure we used to create ground truth keypoint maps. This provides a location map $\bb{n}$.  And we set:
\begin{align}
U_i := -\text{logit} \left(\frac{\|\bb{n} \odot \bb{m}_i\|_F}{ \|\bb{n} \odot \bb{n}\|_F}\right), 
\label{eq:Ui}
\end{align}
where $\|\cdot\|_F$ represents the Frobenius norm,  $\odot$ represents the Hadamard product, and $\text{logit}(x) = \log\left(x/(1-x)\right)$.
Note that since we do not expect a
single placement to explain the entire keypoint location map, we setup the score
as a multiplicative one, with the value only being dependent on the agreement
of the actual keypoints the placement exhibits.

We define the pairwise energy using the GMM model learned from the scene dataset:
\begin{align}
P_{i,j} := -\text{logit}\left(GMM(\delta_\theta^{i,j}, \bb{\delta}_t^{i,j} )\right),
\end{align}
where $\delta_\theta^{i,j}, \bb{\delta}_t^{i,j} $ are the relative orientations and translation of the objects $o_i, o_j$.

We solve for the indicator variables $\{\gamma_i\}$ using OpenGM~\cite{OpenGM} by converting the above formulation into  
a linear program and feeding it to CPLEX~\cite{CPLEX} to find the final set of selected objects.

\vspace{.1in}
\mypara{Refined object fitting}
After selecting the set of objects, the scene mockup is ready. However, we found that our scene priors can also improve the initial object fitting results. To achieve this, we add a term from our GMM model to the regularization term ($L_\text{reg}$) in object fitting. We go through all candidate objects and re-optimize their parameters, keeping the selected objects fixed. 
As noted by Olson et al.~\cite{Olson:2013:IJRR}, the structure of the negative log-likelihood (NLL) of a GMM does not lend itself to
non-linear least squares optimization. Instead, we approximate the NLL of the full GMM by considering it as a Max-Mixture, reducing the NLL to the weighted distance from the closest mixture mean. We define the Max-Mixture likelihood function
\[
 p_\mathrm{Max}(\bb{\delta}) = \max_i w_i N(\bb{\delta} | \bb{\mu}_i, \bb{\Sigma}_i), 
\]
where $\bb{\delta} = \begin{bmatrix} \bb{\delta}_t \\ \delta_\theta
\end{bmatrix}$ is the relative translation and orientation of the new candidate
w.r.t.\ the already placed object, and $w_k$ is the weight of the $k$th mixture in
the model.
We use the sum of negative log-likelihoods of these terms for all selected objects that are within a distance of $\delta_r$ to the refined candidate:
{
\small 
\[ 
-\log(p_\mathrm{Max}(\bb{\delta})) =  \min_k \frac{1}{2} (\bb{\delta} - \bb{\mu}_k)^T \bb{\Sigma}_k^{-1}(\bb{\delta} - \bb{\mu}_k) - \log(w_k\eta_k),\]
}
where $N(\bb{\mu}, \bb{\Sigma})$ represents the normal distribution, and $\eta_k$ is the Gaussian normalization factor for the $k$th mixture. At
optimization time, during each step we find the mixture component $k^*$ that
minimizes this function, and then optimize w.r.t.\ the negative log likelihood
of the Gaussian of that component alone, resulting in the following term to be added to the objective function $L_\text{reg}$ (Equation~\ref{eqn:energy}):
\begin{align}
\frac{1}{2} (\bb{\delta} - \bb{\mu}_{k^*})^T \bb{\Sigma}_{k^*}^{-1}(\bb{\delta} - \bb{\mu}_{k^*}).
\end{align}

\mypara{Refined selection}
Refined candidates and objects selected for the mockup can help in placing additional objects that have subtler cues. Hence, we iterate between refined fitting and refined selection processes. In the refined selection, we assume that previously selected objects cannot be removed, and add the unary term to favor placing new candidates. So, for each candidate placement in the second iteration, we add a term to $U_i$ (Eq.~\ref{eq:Ui}):
{\small 
\begin{align}
-\sum_k \text{logit} (GMM(o_i, o^*_k)^\beta),
\end{align}
}
where $\{o^*_k\}$ are the objects selected at previous iterations. 

% NOTE: hyper parameters (another "implementation details"? go to evaluations?)
% NOTE: data for evaluation => results
% 

%% file: results.tex
\section{Results and Discussion}
\label{sec:ch4:evaluation}

\subsection{Training and test data}
We curated three datasets to evaluate our method. (Datasets to be made available for research use.) 

\mypara{(a) 2D keypoints on indoor images} We downloaded 5000 images from the \textsc{Houzz} website using keywords like living room, kitchen, dining room, meeting room, etc. 
We utilized the Amazon Mechanical Turk platform to obtain keypoints on the images requiring at least 3 workers to agree per image. For each image, we asked the turkers to mark the keypoints of the chairs (maximum of 8 keypoints per chair). Please refer to the supplemental material for details about the web-based annotation interface. 
We convolved these keypoints with a Gaussian filter to simplify the CNN's task of learning of  smooth filters and averaged the results. 

\begin{figure}[h!]
    \centering
    \includegraphics[width=0.99\linewidth]{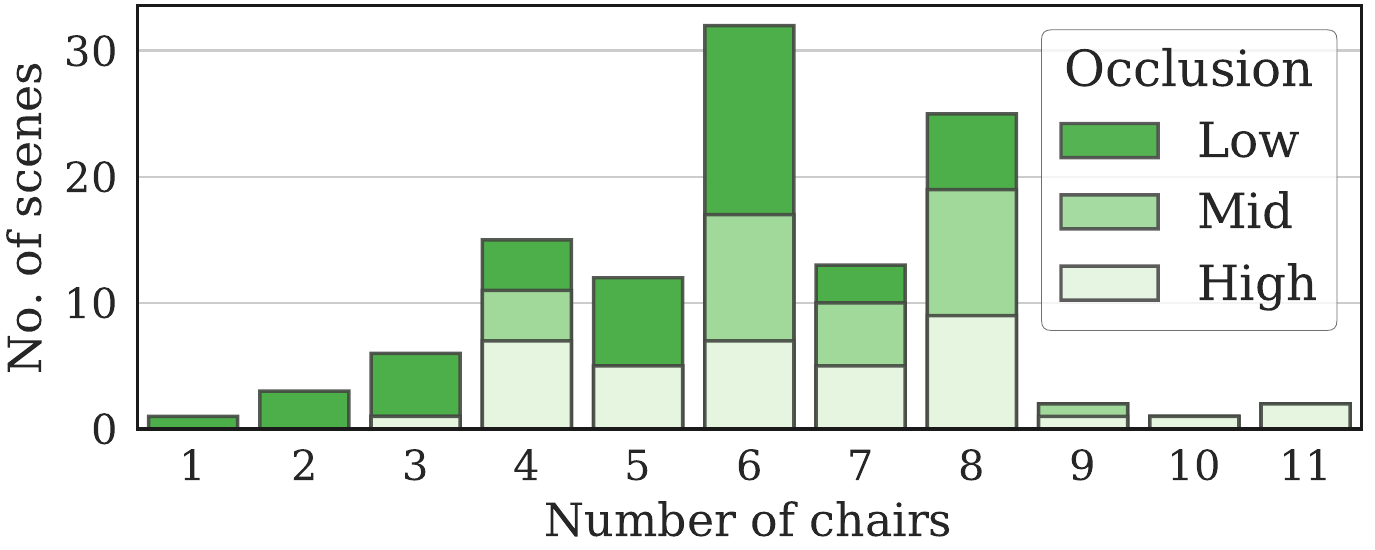}
    \caption{Number of chairs and their estimated visibility distribution in the sampling of images in our annotated \textsc{Houzz} dataset.}
    \label{fig:datasetVisibility}
\end{figure}

\begin{figure*}[t!]
    \centering
    %\def\svgwidth{\linewidth}
    %\import{figures/qualitative_results/}{qualitative_results.pdf_tex}
    \includegraphics[width=\linewidth]{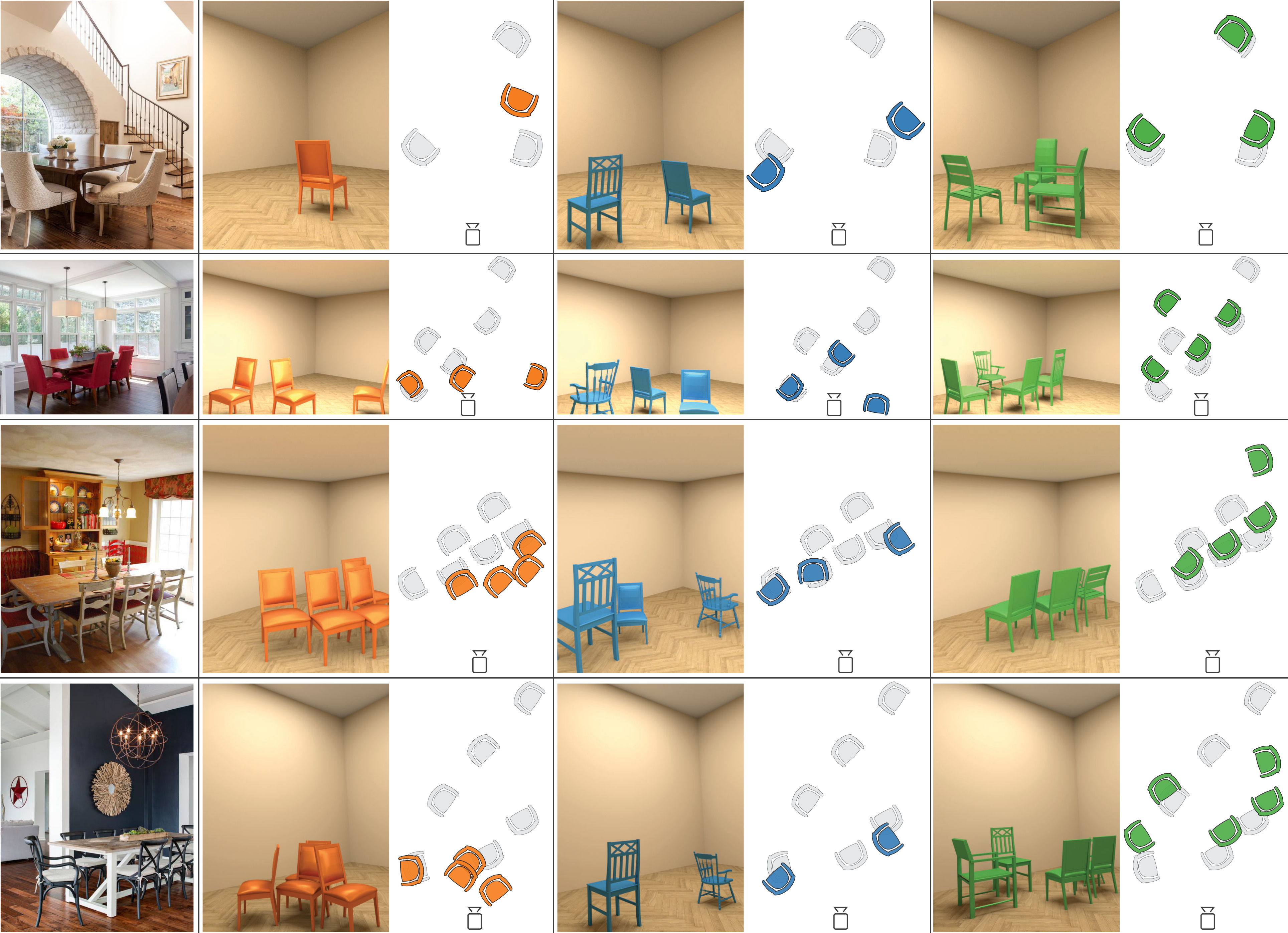}
    \caption{Qualitative comparison of the baseline methods: \textsc{SeeingChairs}~(orange) and \textsc{FasterRCNN3D}~(blue) against \SeeThrough~(green).  Annotated groundtruth poses~(gray) are provided for reference in the top view. Note that our approach both detects ore chairs and correctly aligns them compared to the others. }
    \label{fig:ch4:qualitative_results}
    \vspace{-0.5cm}
\end{figure*}

\mypara{(b) Scene mockup groundtruth}
\label{sec:ch4:ground_truth_annotation}
In order to quantitatively measure the performance of \SeeThrough\ and compare with alternate methods, we require a set of ground truth annotated scenes, i.e., images for
which all the 3D objects (chairs in our case) have been placed manually. We are not aware of a similar dataset with mockups for 3D objects including the (partially) occluded ones.  Hence, we setup another annotation tool in which an object can be placed by clicking and dragging, as well as by annotating a
number of keypoints of the object, and optimizing for its location and scale.
Moreover, objects can be copied and translated along their local coordinate
axes, allowing for quick and precise annotation (see
supplemental for details). We used the automatically estimated camera
parameters for the automatic refinement, while discarding any image with grossly erroneous camera estimates.  We used the tool to  annotate 300 scenes (see Figure~\ref{fig:datasetVisibility}), which were
randomly selected from our \textsc{Houzz} dataset. 
%

%\mypara{(c) 3D models and scenes} For our database models we used the chair models off the ShapeNet~\cite{shapenet2015} database and for the training setup of our network with synthetic  data,  we  used  renders  from  the  PBRS  dataset~\cite{Zhang:2017:CVPR}, which provides 45K houses with 400K high quality renders. We then took the ones containing at least one of the annotated chairs and reprojected the keypoint locations into these renders, yielding one image/keypoint map pair as training data per render,  resulting in a total of $\pm$8000 image/keypoint map pairs.

\mypara{(c) 3D models and scenes} For our database models, we used the chair models from the ShapeNet~\cite{shapenet2015} database and for scene statistics, we used 45K houses from the PBRS  dataset~\cite{Zhang:2017:CVPR}. While the latter comes with 400K physically-based renderings, we tried using these synthetic images to pretrain networks for predicting keypoint maps, but found that fine-tuning a variant of ResNet-50 with weights trained on ImageNet produced more accurate results (see Section~\ref{sec:ch4:discussion} for more details).

\subsection{Performance Measures and Parameters}

\vspace*{0.1in}
\mypara{Hyperparameters}
Our optimization pipeline depends on a number of parameters that we optimized using HyperOpt~\cite{HyperOpt}, which employs a Tree of Parzen Estimators~\cite{Bergstra:2013:ICML}.
We used the \PctCorrFull\ measure as our objective measure. 
As ground truth data, we used 10 scenes fully annotated specifically for this purpose,
in the same way as the data used for evaluation (see above).
%See Table~\ref{tab:ch4:hyperparameters} for a list of resulting hyper parameter values.
See supplemental material for the list of resulting hyper parameter values.

\begin{figure*}[t!]
    \centering
    \includegraphics[width=\textwidth]{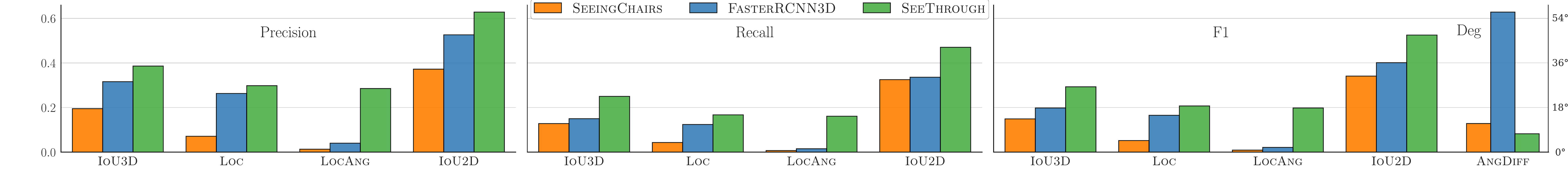}
    \caption{Quantitative performance of \SeeThrough\ against the state-of-the-art method-based baseline methods. We outperform the baselines significantly across all the measures. Please refer to supplemental for the tabulated values. }
    \label{fig:quantMeasures_all}
\end{figure*}

\mypara{Quantitative measures}
We use \emph{source} and \emph{target}
to denote the two scenes between which a measure is computed. We
specifically do not use `result scene' and `ground truth scene' as
the ground truth acts as a target to compute precision, and acts as source 
to compute recall.
%they can act as either source or target scene in most measures. 

We denote
the objects in the source and target scene as $o_S \in \bb{S}$, $o_T \in \bb{T}$, 
respectively. We use $J_3(o_S, o_T)$ and $J_2(o_S, o_T)$ to represent the Jaccard index
or \emph{intersection-over-union} (IoU) of the bounding boxes of $o_S$ and
$o_T$ in 3D world space and 2D screen space, respectively. Finally, given an
object $o_S$ we define the `$J_i^*$ correspondence' with $\bb{T}$ as the object with the MaxIoU with $o_S$ as: $ J_i^*(o_S, \bb{T}) := \arg\max_{o_T \in \bb{T}}
J_i(o_S, o_T). $
Intuitively, this returns, for a given object, the {\em best matching} object from
the other scene in terms of overlap. Next, we briefly describe our selected measures (see supplemental for details).

{\noindent  (a)~\AvgMaxIou:} This measures average IoU for 3D bounding boxes around objects. 
Specifically, given a source scene and a target scene, we average MaxIoU across all objects in the source scene 
(measuring IoU overlap with the corresponding object in the target). 

%This measure takes a source scene and a target
        %scene, and records the accuracy with which the volumes of the objects
%        in the source scene agree with the objects in the target scene.
        %Specifically, for each object in the source scene, we record the IoU of
        %the object with its MaxIoU correspondence.  This measure is averaged
        %over all objects in the source scene to produce the final measure.

{\noindent (b)~\AvgMaxDIoU:} Similar to \AvgMaxIou, this measure averages IoU for 2D bounding boxes around projected objects. 
        
{\noindent (c)~\PctCorrLoc:} This measures the fraction of correct locations of objects in the source scene 
        with respect to the target. We consider every object in the source scene that has a $J_3^*$ correspondence over a  threshold
        $\tau_{J}$ to have a correct location.

{\noindent (d)~\PctCorrFull:} Similar to \PctCorrLoc, this measures additionally requires the angle difference to be under a threshold $\tau_{\theta}$.

{\noindent (e)~\AngDiff:} This measures the average angle difference for the objects that have a
        correct location. %This measure is symmetrical.

%This measures the average maximum IoU of the
%        bounding boxes of each projected object in the source scene with the
%        bounding boxes of the projected objects in the target scene. 

\subsection{Baselines: State-of-the-art Alternatives}
\label{sec:ch4:baselines}
We are not aware of prior research focusing on producing scene mockups in the presence of {\em significant occlusion}. Hence, we created two baselines by combining relevant state-of-the-art methods. We convert the output
of each baseline (in both cases 3D pose but 2D image space locations of chairs) to
our comparable 3D scene mockup format.

\mypara{(a) \seeingChairs} Aubry et al.~\cite{Aubry:2014:CVPR} proposed a method to find chairs by
matching so-called `discriminative visual elements' (DVE) from a set of
rendered views of 1000+ chair models with any input image. These DVEs are
linear classifiers over HOG features \cite{Dalal:2005:CVPR} learned from the
rendered views in a discriminative fashion. 
At training time, they are learned at multiple scales while keeping only the most discriminative ones for matching. At test time,
a patch-wise matching process finds the best-matching image and rendered patch pairs,
and then finds sets of pairs that come from the same rendered view (see \cite{Aubry:2014:CVPR} for details).

The above method outputs scored image space bounding boxes together with a specific
chair model and pose. For our 3D
performance measures, however, we need the
output in the form of a 3D scene. Hence, we convert each set of bounding
box, pose, and chair model to a 3D scene. Using our estimated camera, we optimize the location (in the
xz-plane) of the 3D model without changing its pose, such that the 2D bounding
box of the projected model matches as closely as possible with the detected
bounding box using a least-squares formulation (solved using  Ceres~\cite{Ceres}).

\mypara{(b) \fasterRCNN} As the second baseline, we combine a convolutional neural network (CNN) trained for image-space 
object detection and another CNN trained for 3D object interpretation. Specifically, we use FasterRCNN~\cite{Ren:2015:NIPS} to extract bounding boxes of chairs from the input
image and then feed these regions of interest to 3D-INN~\cite{wu2016single}, which produces a
templated chair model consisting of a set of predefined 3D keypoints as well as
a pose estimate. 
Since our set of keypoints is a subset of the keypoints
produced by 3D-INN, we use our 3D candidate generation part of \SeeThrough\  
to convert the extracted keypoints to a 3D chair for the resultant scene mockup.

\subsection{Evaluation and Discussion}
\label{sec:ch4:discussion}

We ran \SeeThrough\ and the two baseline methods on the full ground truth
annotated scene set (Section~\ref{sec:ch4:ground_truth_annotation}). A sampling
of results can be seen in Figure~\ref{fig:ch4:qualitative_results}. (Further  visualization
for  100 scenes in our groundtruth set can be found in the supplementary material.) 

The baseline methods perform well when there is no occlusion in the scene.
Specifically, chairs that are clearly visible are reconstructed reliably as the direct visual
information is sufficient to make an accurate 
inference about the objects' pose and identity. However, when chairs are partly
occluded, the methods break down quickly. In contrast, \SeeThrough, by incorporating co-occurrence object model,  is more
often able to recover from these situations. 

\begin{figure}[h!]
   \centering
    \includegraphics[width=\linewidth]{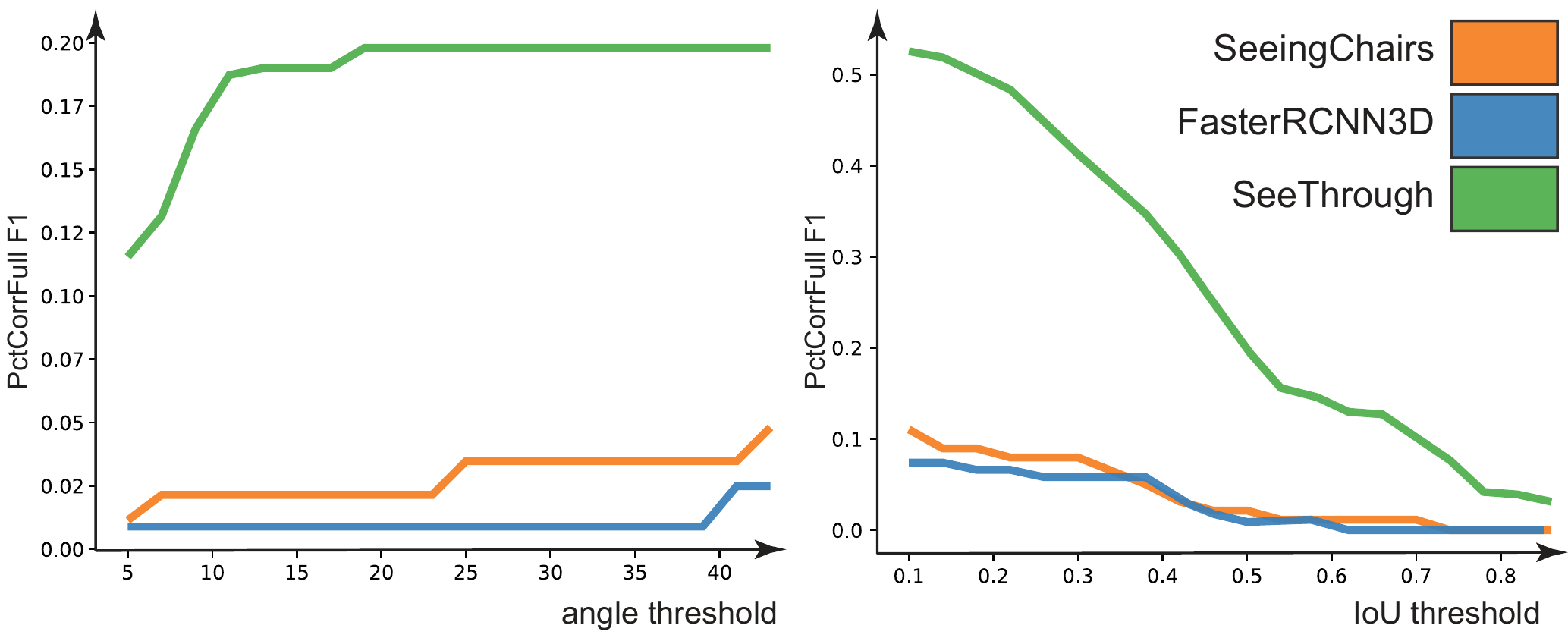}
    \caption{Performance variation according to \PctCorrFull\ F1 measure for \SeeThrough\ and the two baseline methods under varying angle and IoU thresholds. We perform significantly better across both the threshold ranges.  }
    \label{fig:ch4:performance_changes}
\end{figure}

This difference in performance is also reflected in the quantitative results (see Figure~\ref{fig:quantMeasures_all}). Our method outperforms the baselines on all
counts.  Additionally, in Figure~\ref{fig:ch4:performance_changes}, we show how the
\PctCorrFull\ measure changes under varying thresholds of angle ($\tau_\theta$) and IoU ($\tau_J$).

\mypara{Performance under increasing occlusion} In order to specifically test performance under varying occlusion, we sorted the groundtruth annotated \textsc{Houzz} dataset into categories based on the extent of the visible chairs.
We approximate visibility as follows: we compute how many chairs lie along view rays connecting the estimated camera location with points on a discrete grid on the image plane. We used the objects' bounding boxes for this visibility computation. Higher values denote more occlusion (as there are more chairs along the view rays). Figure~\ref{fig:teaser} shows that while all the three methods perform comparably under low occlusion, only \SeeThrough\ continues to have a high success rate under medium to heavy occlusion.

\mypara{Effect of multiple iterations} In Section~\ref{sec:ch4:ablation}, we demonstrate the positive utility of multiple iterations to \SeeThrough. 
One of our key observations is that high-confidence objects (e.g., unoccluded objects) are easier to detect, and hence can provide valuable contextual information in reinforcing the weaker signals (e.g., partially occluded objects). This behavior results in higher detection rates using iterations and believed to be also functional in the human perception systems~\cite{brainWorks12,brainWorks17}.

\mypara{Utility of synthetic data} 
We found that training on synthetic datasets~\cite{Zhang:2017:CVPR} for predicting image-space keypoint maps led to unsatisfactory results. 
For this experiment, we took all renderings from 400K images that contain at least one of the annotated chairs and reprojected the keypoint locations from corresponding 3D models into these renders, yielding one image/keypoint map pair as training data per render,  resulting in a total of 8000 image/keypoint map pairs.
We experimented with three different training setups: 
(i)~network trained with only synthetic data;  (ii)~network first trained with synthetic data, 
and then refined using real data, and (iii)~network trained with only real data. 

The best performance on the test set resulted from setup \#iii, i.e.
training with only real data. %Apparently, the shortcomings of the synthetic
%data mentioned above were of higher importance than expected. 
One likely explanation is that training the network with the synthetic data first
steers away the network weights from those that were the result of the ImageNet
pretraining, which already encompass a high general understanding of real
photographs. 
%The numbers show that this initial information is more valuable than the
%extent of the synthetic data as well as its structural similarity to our test
%data.

\begin{figure}[t]
    \centering
    \includegraphics[width=\linewidth]{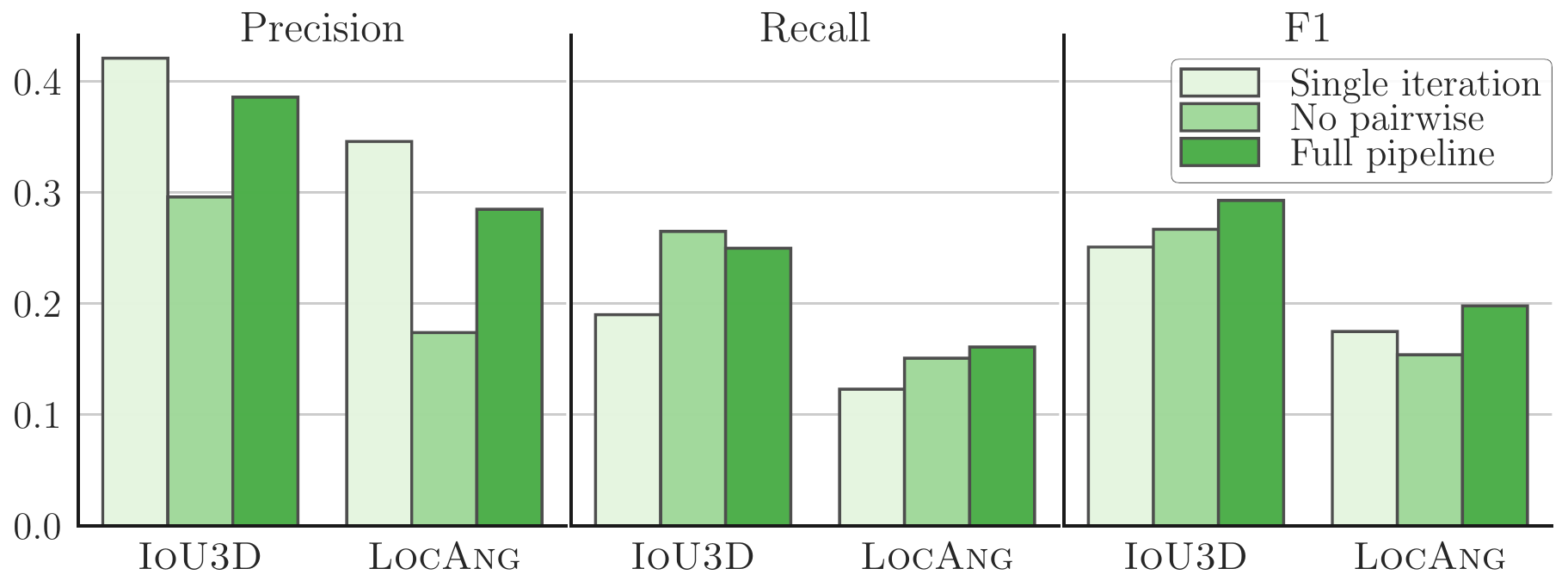}
    \caption{Ablation study evaluating the importance of the different stages of \SeeThrough.}
    \label{fig:ablation}
\end{figure}

\subsection{Ablation Study}
\label{sec:ch4:ablation}
We evaluated the importance of the individual steps of \SeeThrough\  to the final performance (see Figure~\ref{fig:ablation} and supplemental). Specifically,
we ran our pipeline on the full test set under two weakening conditions:
(a)~we disable all pairwise
costs and run the remaining pipeline based solely on the keypoint location maps; and 
(b)~we disable iterations by running 
the second and third stage only once, thus removing the possibility of the candidate generation stage benefiting from previously placed objects. 

\mypara{Discussion} Although \AvgMaxDIoU\ recall increases when disabling scene statistics (option \#a), the precision goes down significantly. This is true  
as the pairwise costs by themselves do not propose new objects -- they only make output mockups more precise by pruning objects
that do not agree with others. 
In contrast, using only a single iteration (option \#b) increases precision, but recall takes a significant hit. This is not surprising, 
as in the later iterations the keypoint location maps have decreased influence relative to the pairwise costs. As a result,  while objects with weaker
keypoint response are more easily found, false positives also become more likely. Overall, the combined \AvgMaxDIoU\ F1 measure is highest
for the full \SeeThrough\ as well as the \PctCorrFull\ F1 measure.

%% file: conclusion.tex
\section{Conclusion}
We proposed \SeeThrough, a method for automatically finding partially occluded chairs in a photograph of a structured scene.
Our key insight is the incorporation of higher level
scene statistics that allows more accurate reasoning in scenes containing medium to high levels of occlusion. 
We demonstrate considerable quantitative and qualitative performance improvements across multiple measures. 

Our method suffers from limitations that suggest a number of future research directions. First, we plan to extend the evaluation to a more expansive class of objects beyond chairs.  Second, we think exploring templates that can express a broader understanding of the multi-object spatial relationships is a promising future direction.

%% file: seeThrough.bbl
\begin{thebibliography}{10}\itemsep=-1pt

\bibitem{CPLEX}
{IBM ILOG CPLEX Optimizer}.
\newblock \url{https://www.gams.com/latest/docs/S_CPLEX.html}.

\bibitem{Ceres}
S.~Agarwal, K.~Mierle, and Others.
\newblock Ceres solver.
\newblock \url{http://ceres-solver.org}.

\bibitem{OpenGM}
B.~Andres, T.~Beier, and J.~Kappes.
\newblock {OpenGM}: {A} {C++} library for discrete graphical models.
\newblock {\em CoRR}, abs/1206.0111, 2012.

\bibitem{Aubry:2014:CVPR}
M.~Aubry, D.~Maturana, A.~A. Efros, B.~C. Russell, and J.~Sivic.
\newblock Seeing 3d chairs: Exemplar part-based 2d-3d alignment using a large
  dataset of {CAD} models.
\newblock In {\em Proc. {IEEE CVPR}}, pages 3762--3769, 2014.

\bibitem{Bergstra:2013:ICML}
J.~Bergstra, D.~Yamins, and D.~Cox.
\newblock Making a science of model search: Hyperparameter optimization in
  hundreds of dimensions for vision architectures.
\newblock In {\em Proc. {ICML}}, pages 115--123, 2013.

\bibitem{shapenet2015}
A.~X. Chang, T.~Funkhouser, L.~Guibas, P.~Hanrahan, Q.~Huang, Z.~Li,
  S.~Savarese, M.~Savva, S.~Song, H.~Su, J.~Xiao, L.~Yi, and F.~Yu.
\newblock {ShapeNet: An Information-Rich 3D Model Repository}.
\newblock Technical Report arXiv:1512.03012 [cs.GR].

\bibitem{choi2015indoor}
W.~Choi, Y.-W. Chao, C.~Pantofaru, and S.~Savarese.
\newblock Indoor scene understanding with geometric and semantic contexts.
\newblock {\em IJCV}, 112(2):204--220, 2015.

\bibitem{Dalal:2005:CVPR}
N.~Dalal and B.~Triggs.
\newblock Histograms of oriented gradients for human detection.
\newblock In {\em Proc. {IEEE CVPR}}, pages 886--893, 2005.

\bibitem{dasgupta2016delay}
S.~Dasgupta, K.~Fang, K.~Chen, and S.~Savarese.
\newblock Delay: Robust spatial layout estimation for cluttered indoor scenes.
\newblock In {\em Proc. {IEEE CVPR}}, pages 616--624, 2016.

\bibitem{del2012bayesian}
L.~Del~Pero, J.~Bowdish, D.~Fried, B.~Kermgard, E.~Hartley, and K.~Barnard.
\newblock Bayesian geometric modeling of indoor scenes.
\newblock In {\em Proc. {IEEE CVPR}}, pages 2719--2726. IEEE, 2012.

\bibitem{brainWorks12}
J.~J. DiCarlo, D.~Zoccolan, and N.~C. Rust.
\newblock How does the brain solve visual object recognition?
\newblock {\em Neuron}, 73(3):415--434, 2012.

\bibitem{Fisher:2012:SIGGASIA}
M.~Fisher, D.~Ritchie, M.~Savva, T.~Funkhouser, and P.~Hanrahan.
\newblock Example-based synthesis of 3d object arrangements.
\newblock {\em Proc. ACM/SIGGRAPH Asia}, 2012.

\bibitem{Fisher:2015:SIGGRAPH}
M.~Fisher, M.~Savva, Y.~Li, P.~Hanrahan, and M.~Nie{\ss}ner.
\newblock Activity-centric scene synthesis for functional 3d scene modeling.
\newblock {\em Proc. ACM/SIGGRAPH}, 34(6):179, 2015.

\bibitem{brainWorks17}
A.~M. Fyall, Y.~El-Shamayleh, H.~Choi, E.~Shea-Brown, , and A.~Pasupathy.
\newblock Dynamic representation of partially occluded objects in primate
  prefrontal and visual cortex.
\newblock {\em eLife}, 2017.

\bibitem{He:2016:CVPR}
K.~He, X.~Zhang, S.~Ren, and J.~Sun.
\newblock Deep residual learning for image recognition.
\newblock In {\em Proc. {IEEE CVPR}}, pages 770--778, 2016.

\bibitem{hedau2009recovering}
V.~Hedau, D.~Hoiem, and D.~Forsyth.
\newblock Recovering the spatial layout of cluttered rooms.
\newblock In {\em Proc. {ICCV}}, pages 1849--1856. IEEE, 2009.

\bibitem{Hedau:2009:ICCV}
V.~Hedau, D.~Hoiem, and D.~A. Forsyth.
\newblock Recovering the spatial layout of cluttered rooms.
\newblock In {\em Proc. {ICCV}}, pages 1849--1856, 2009.

\bibitem{hoiem2005automatic}
D.~Hoiem, A.~A. Efros, and M.~Hebert.
\newblock Automatic photo pop-up.
\newblock {\em tog}, 24(3):577--584, 2005.

\bibitem{huang2015single}
Q.~Huang, H.~Wang, and V.~Koltun.
\newblock Single-view reconstruction via joint analysis of image and shape
  collections.
\newblock {\em ACM Transactions on Graphics (TOG)}, 34(4):87, 2015.

\bibitem{HyperOpt}
HyperOpt.
\newblock {HyperOpt}, 2017.

\bibitem{Izadinia:2016:Arxiv}
H.~Izadinia, Q.~Shan, and S.~M. Seitz.
\newblock {IM2CAD}.
\newblock {\em CoRR}, abs/1608.05137, 2016.

\bibitem{izadinia2017im2cad}
H.~Izadinia, Q.~Shan, and S.~M. Seitz.
\newblock Im2cad.
\newblock In {\em Proc. {IEEE CVPR}}, 2017.

\bibitem{kholgade20143d}
N.~Kholgade, T.~Simon, A.~Efros, and Y.~Sheikh.
\newblock 3d object manipulation in a single photograph using stock 3d models.
\newblock {\em ACM Transactions on Graphics (TOG)}, 33(4):127, 2014.

\bibitem{kmyg_acquireIndoor_sigga12}
Y.~M. Kim, N.~J. Mitra, D.-M. Yan, and L.~Guibas.
\newblock Acquiring 3d indoor environments with variability and repetition.
\newblock {\em ACM Transactions on Graphics}, 31(6):138:1--138:11, 2012.

\bibitem{Li:2015:CGF}
Y.~Li, A.~Dai, L.~J. Guibas, and M.~Nie{\ss}ner.
\newblock Database-assisted object retrieval for real-time 3d reconstruction.
\newblock {\em Comput. Graph. Forum}, 34(2):435--446, 2015.

\bibitem{lim2014fpm}
J.~J. Lim, A.~Khosla, and A.~Torralba.
\newblock Fpm: Fine pose parts-based model with 3d cad models.
\newblock In {\em Proc. {ECCV}}, pages 478--493. Springer, 2014.

\bibitem{Lim:2013:ICCV}
J.~J. Lim, H.~Pirsiavash, and A.~Torralba.
\newblock Parsing {IKEA} objects: Fine pose estimation.
\newblock In {\em Proc. {ICCV}}, pages 2992--2999, 2013.

\bibitem{mallya2015learning}
A.~Mallya and S.~Lazebnik.
\newblock Learning informative edge maps for indoor scene layout prediction.
\newblock In {\em Proc. {ICCV}}, pages 936--944, 2015.

\bibitem{Mattausch:2014:CGF}
O.~Mattausch, D.~Panozzo, C.~Mura, O.~Sorkine{-}Hornung, and R.~Pajarola.
\newblock Object detection and classification from large-scale cluttered indoor
  scans.
\newblock {\em Comput. Graph. Forum}, 33(2):11--21, 2014.

\bibitem{Monszpart:2015:SIGGRAPH}
A.~Monszpart, N.~Mellado, G.~J. Brostow, and N.~J. Mitra.
\newblock Rapter: rebuilding man-made scenes with regular arrangements of
  planes.
\newblock {\em Proc. ACM/SIGGRAPH}, 34(4):103, 2015.

\bibitem{Olson:2013:IJRR}
E.~Olson and P.~Agarwal.
\newblock Inference on networks of mixtures for robust robot mapping.
\newblock {\em I. J. Robotics Res.}, 32(7):826--840, 2013.

\bibitem{Ren:2015:NIPS}
S.~Ren, K.~He, R.~B. Girshick, and J.~Sun.
\newblock Faster {R-CNN:} towards real-time object detection with region
  proposal networks.
\newblock In {\em Proc. {NIPS}}, pages 91--99, 2015.

\bibitem{schwing2013box}
A.~G. Schwing, S.~Fidler, M.~Pollefeys, and R.~Urtasun.
\newblock Box in the box: Joint 3d layout and object reasoning from single
  images.
\newblock In {\em Proc. {ICCV}}, pages 353--360, 2013.

\bibitem{Shao:2014:SIGGRAPH}
T.~Shao, A.~Monszpart, Y.~Zheng, B.~Koo, W.~Xu, K.~Zhou, and N.~J. Mitra.
\newblock Imagining the unseen: stability-based cuboid arrangements for scene
  understanding.
\newblock {\em Proc. ACM/SIGGRAPH}, 33(6):209:1--209:11, 2014.

\bibitem{tulsiani2015viewpoints}
S.~Tulsiani and J.~Malik.
\newblock Viewpoints and keypoints.
\newblock In {\em Proc. {IEEE CVPR}}, pages 1510--1519, 2015.

\bibitem{Wei:2016:CVPR}
S.~Wei, V.~Ramakrishna, T.~Kanade, and Y.~Sheikh.
\newblock Convolutional pose machines.
\newblock In {\em Proc. {IEEE CVPR}}, pages 4724--4732, 2016.

\bibitem{wu2016single}
J.~Wu, T.~Xue, J.~J. Lim, Y.~Tian, J.~B. Tenenbaum, A.~Torralba, and W.~T.
  Freeman.
\newblock Single image 3d interpreter network.
\newblock In {\em Proc. {ECCV}}, pages 365--382. Springer, 2016.

\bibitem{xiao2012localizing}
J.~Xiao, B.~Russell, and A.~Torralba.
\newblock Localizing 3d cuboids in single-view images.
\newblock In {\em Proc. {NIPS}}, pages 746--754, 2012.

\bibitem{zhang2014panocontext}
Y.~Zhang, S.~Song, P.~Tan, and J.~Xiao.
\newblock Panocontext: A whole-room 3d context model for panoramic scene
  understanding.
\newblock In {\em Proc. {ECCV}}, pages 668--686. Springer, 2014.

\bibitem{Zhang:2017:CVPR}
Y.~Zhang, S.~Song, E.~Yumer, M.~Savva, J.~Lee, H.~Jin, and T.~A. Funkhouser.
\newblock Physically-based rendering for indoor scene understanding using
  convolutional neural networks.
\newblock In {\em Proc. {IEEE CVPR}}, 2017.

\end{thebibliography}
